\pdfoutput=1

\documentclass[fleqn,10pt]{wlscirep}
\usepackage[utf8]{inputenc}
\usepackage[T1]{fontenc}

\title{Learning Bone Suppression from Dual Energy Chest X-rays using Adversarial Networks}

\author[1]{Dong Yul Oh}
\author[2,*]{Il Dong Yun}
\affil[1]{Interdisciplinary Program in Bioengineering, Seoul National University, Korea}
\affil[2]{Division of Computer and Electronic System Engineering, Hankuk University of Foreign Studies, Korea}
\affil[*]{Correspondence: yun@hufs.ac.kr}


\begin{abstract}
Suppressing bones on chest X-rays such as ribs and clavicle is often expected to improve pathologies classification. These bones can interfere with a broad range of diagnostic tasks on pulmonary disease except for musculoskeletal system. Current conventional method for acquisition of bone suppressed X-rays is dual energy imaging, which captures two radiographs at a very short interval with different energy levels; however, the patient is exposed to radiation twice and the artifacts arise due to heartbeats between two shots. In this paper, we introduce a deep generative model trained to predict bone suppressed images on single energy chest X-rays, analyzing a finite set of previously acquired dual energy chest X-rays. Since the relatively small amount of data is available, such approach relies on the methodology maximizing the data utilization. Here we integrate the following two approaches. First, we use a conditional generative adversarial network that complements the traditional regression method minimizing the pairwise image difference. Second, we use Haar 2D wavelet decomposition to offer a perceptual guideline in frequency details to allow the model to converge quickly and efficiently. As a result, we achieve state-of-the-art performance on bone suppression as compared to the existing approaches with dual energy chest X-rays.
\end{abstract}

\begin{document}

\flushbottom
\maketitle

\thispagestyle{empty}


\section{Introduction}
Over twenty-thousand people die every year due to diseases related to the lung and its surroundings, such as chronic obstructive pulmonary disease (COPD), emphysema, and pneumonia \cite{1}. Radiologists first obtain chest X-rays in order to diagnose these pulmonary diseases, however, the ribs interfere with careful observation of the lesions, which frequently occurs near parenchyma, heart, peritoneum, etc. except for musculoskeletal system. Previous studies by \cite{2,3} have proved that lung cancer lesions located behind ribs potentially have key features associated with abnormalities. In addition, most patients, particularly those who need regular observation, are able to cope with more precise pathologic outcomes through the difference between the current image and the one previously recorded. The process for matching two images is required but the ribs could also disturb the diagnosis.

Currently, the commercialized method for acquisition of bone suppressed X-rays is dual energy imaging \cite{4}, which captures two radiographs at a very short interval with different energy levels. It performs bone cancellation by exploiting subtraction between the attenuation of soft tissue and bone at different intensities. However, this method has a significant clinical defect in which the patient is exposed to the radiation twice and artifacts arise due to heart beat between two shots. Although low-dose imaging techniques have been developed, it is rarely true that X-ray exposure does not increase the probability of causing other diseases such as skin cancer. Since heart beat is not a function that a human can temporarily stop, additional techniques are required to solve the artifacts caused by the heart movement. Furthermore, a specialized equipment, which is expensive to purchase and maintain, is required to obtain dual energy X-rays (DXRs). Other conventional techniques are limited in their performance because X-rays, technically radiographs, have a wide range of clinical settings in medical imaging, and inter-class variation is very high.

We therefore tackle this problem with a novel approach using deep learning based model to learn bone suppression on single energy chest X-rays from previously acquired dual energy chest X-rays. Similar problems have already been addressed by \cite{5,6,7,8}. As big data become readily available, most solutions adopt the architectures of such approach as existing family of convolutional auto-encoders \cite{9}. They have optimized the network parameters to minimize the \textit{average} pixel-wise difference (with some other designed pixel-related functions) between the prediction and its ground truth. This is very straightforward and easy for the model to converge, however the bone suppressed images are quite blurry due to the nature of minimizing average pixel values, which we will discuss by comparing with our approach in Section 4.1 and 4.3.

Inspired by the recent success of the deep generative models \cite{10,11,12,13}, we fundamentally focus not only on de-noising approach that considers bone as a noise but also learning conditional probability distribution of bone suppressed image respect to its original one. The approach of \cite{12} is the closest to ours in using Generative Adversarial Networks (GANs) \cite{14}. The objective function to optimize the model parameters is the amount of noise, Euclidean distance between pairwise outputs and labels, which is equivalent to other previous approaches. Here we add an adversarial training framework to maintain the sharpness of specific lesions on single energy X-rays and avoid undesirably suppressing them. The key difference from \cite{12} is the choice of improved techniques to leverage a finite set of data based on the original GAN framework.

\subsection{Main Contributions}
This work first of all addresses the problem of minimizing average pixel-wise differences to learn bone suppression on single energy chest X-rays. Existing conditional adversarial networks of \cite{12} is purposely modified to accomplish such a goal. Our contributions are summarized as:
\begin{itemize}
\item This work experimentally verifies that adversarial training framework for modeling de-noising approach with conditional image-to-image translation on bone suppression is able to outperform existing state-of-the-art methods.
\item We propose to explicitly exploit frequency details using Haar 2D wavelet decomposition to offer a perceptual guideline for minimizing pairwise image differences.
\item To the best of our knowledge, the model discussed in this paper is the first approach using deep generative models for bone suppression with DXRs, which has been rigorously evaluated.
\end{itemize}

\subsection{Related Work}
The present work is a partial solution of bone suppression on chest X-rays improving pathologic outcomes of both computer-assisted diagnosis (CAD) and radiologists. Many recent efforts to address this problem have been proposed. All of them utilize their method to extract specific information of bones from given chest X-rays and recognize where to suppress.

Bone suppression was first introduced by \cite{15}, removing the dominant effects of the bony structure within the X-ray projection and reconstructing residual soft tissues components. Most of general studies in relation to bone suppression received relatively less attention and have been conducted for very specific purpose until the actual clinical effect from bone suppression has been verified. However, \cite{2} proved that currently learned diagnosis suffers from lung cancer lesions obscured by anatomical structures such as ribs, and \cite{3} showed that the superposition of ribs highly affects the performance of automatic lung cancer detection. Both studies re-examined the invisibility of abnormalities caused by the superposition of bones and the improvement of automatic or human-level pathologic classification by the detection of these abnormalities.

Since then, great progress has been done in bone suppression. We categorize them into deep learning and non-deep learning approaches. For non-deep learning approaches, one of the most sensational method that received much attention in medical fields is dual energy imaging \cite{4}. It also refers to dual energy subtraction (DES) since it acquires information about specific intensities through a series of subtractions between two X-rays at different energies. Both images at different energies have different attenuation values, hence they can be subtracted to perform bone or tissue cancellation that is able to detect the lesion such as a calcified nodule that did not appeared in either of them. \cite{16} employed Active Shape Model, which is a parametric model of a curve for bones where the parameters are determined from the statistics of many sets of points in similar images, then the segmentation data is used to remove bones by subtraction. \cite{17} followed a similar curve fitting model to get rib segments obtained through Gabor filtering, and used several pre-processing from CAD, local contrast enhancement and lung segmentation. \cite{18} refined the final ribs with the dynamic programming-based active contour algorithm. The key aspects of these previous methods are detecting the position of lung and ribs border first and finally refining the final rib shadows based on vertical intensity profiles.

As deep learning algorithms are further developed, current related studies focus more on deep learning based model on bone suppression. \cite{5} used a massive artificial neural network, which the sub regions of input passes linear dense layers with single output, to obtain the bone image from a single energy chest X-ray. Then they subtract the bone image from the original image to yield virtual dual energy image, similar to a soft-tissue image. \cite{6}. the extension model of \cite{5}, additionally employed a total variation-minimization smoothing method and multiple anatomically specific networks to improve previously achieved performance. A new approach combined with deep learning and dual energy X-rays data has been commonly used recently; \cite{7} trained with 404 dual-energy chest X-rays with a multi-scale approach, and also subtracted the bone image from the original image to obtain a virtual soft tissue image using its vertical gradient as previously introduced. \cite{8} proposed two end-to-end architecture, convolutional auto-encoder network and non-down-sampling convolutional network that directly output the bone suppressed images based on DXR training set. They combined mean squared error (MSE) with the structural similarity index (SSIM) that addresses sensitivity of the human visual system to changes in local structure \cite{19}.

Such a naive adoption of convolutional auto-encoder families often fails to capture the sharpness since the network misses high frequency details, which are the main reason of blurry images, in its encoding and decoding system. \cite{9} have overcome this limitation and achieved high performance on segmentation task with skip connection in the auto-encoding process. The segmentation task can be addressed by creating mask with its pixel-wise probability, however, with an intensity profile in the bone suppression task can potentially act as a bias. \cite{12} employed very heuristic loss function using conditional GAN framework for image translation similarly to neural style transfer. The success of such approach motivates us to do research on more effective and easier method not only to converge on learning bone suppression from a finite set of DXRs, also eliminate bias in suppressing region. We combine the suppressing noisy bones approach with image-to-image translation and purposely re-designed existing conditional adversarial network; the input system and improved techniques in the training process.

\section{Background}

\subsection{Generative Adversarial Networks}
This study aims to learn bone suppression on single energy X-rays from previously acquired DXRs through de-noising approach with conditional image-to-image translation. We use adversarial training within GAN framework \cite{14} to learn the conditional probability distribution of the output (bone suppressed X-ray images) according to the input (original X-ray images). 

\begin{figure}[h]
\begin{center}
\includegraphics[width=5.5cm]{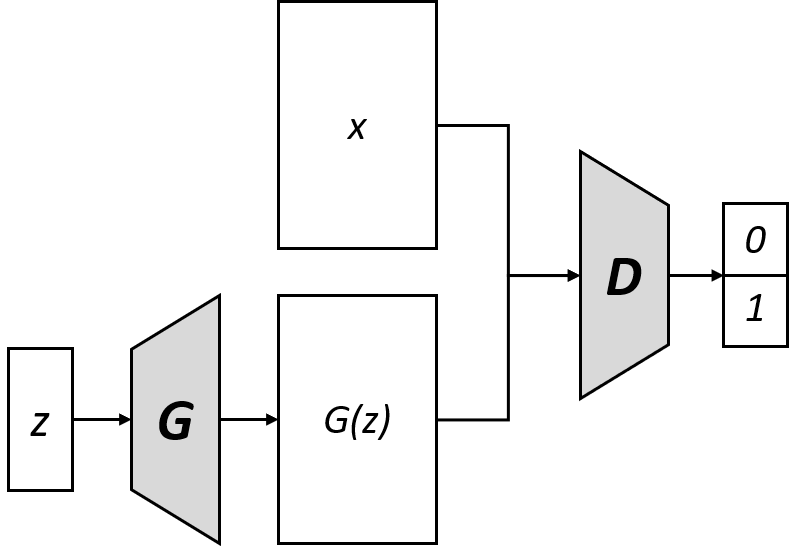}
\end{center}
\caption{The overall schematic of Generative Adversarial Networks.}
\end{figure}

GAN is a generative model that consists of two networks called \textit{generator} and \textit{discriminator }in an adversarial relationship. The generator creates an image similar to the training set, and the discriminator distinguish whether the input is a fake image, which comes from the generator, or a real one coming from the training set. As depicted in Figure 1, the GAN is a structured probabilistic model. The generator is a differentiable function $G$, which basically takes latent variable $z$ for the prior information of the model, then outputs the samples $G(z)$ that are intended to be drawn from the same distribution as observed variables $x$. Here $z$ is regarded as random noise of which sampling method is generally taken in commonly known distribution such as Gaussian or exponential family. The discriminator is a differentiable function $D$ which is a binary classifier taking both $x$ and $G(z)$ and outputs a single probability for either case, $D(x)$ or $D(G(z))$. The discriminator thereby is trained with two mini-batch datasets for real and fake samples unlike the usual case in traditional supervised learning. In this scenario, two networks compete; the discriminator strives to make $D(x)$ to be near 1 while $D(G(z))$ to 0, which can be derived from binary cross-entropy using sigmoid function. Thus, the cost function of the discriminator is as follows:

\begin{equation}
J^{(D)}(\theta^{(D)},\theta^{(G)}) = -\frac{1}{2}\mathbb{E}_{x\sim p_{data}}\log D(x) - \frac{1}{2}\mathbb{E}_{z \sim p_{z}}\log (1-D(G(z)))
\end{equation}

where $\theta^{(D)}$ and $\theta^{(G)}$ are the parameter of generator and discriminator, respectively. (1) offers extremely huge penalty if the discriminator does not properly distinguish both cases. This algorithm often refers to the \textit{game theory} competing the participants (players), where the player's cost is dependent each other and each player cannot control the other player's parameters, hence GAN framework is called adversarial training. The simplest solution is a \textit{Nash equilibrium} corresponding to the $G(z)$ being drawn from the same distribution as the training data $x$, and $D(x) = 0.5$ for all $x$ in this scenario. This is also regarded as a zero-sum game or minimax game that the goal is for the sum of the players' cost is to be zero. Therefore, the cost function for the generator is:

\begin{equation}
J^{(G)} = -J^{(D)}
\end{equation}

However, this minimax game algorithm is very inefficient in an actual training process. Minimizing cross-entropy has been proven for its efficiency because the loss never saturates when the network fails to predict given problem. (2) intuitively shows that when the discriminator minimizes its cross-entropy, the generator maximizes the same cross-entropy. In other words, the gradient vanishing problem where the gradient saturates to 0, occurs in the generator and vice-versa. To end this, we maintain the concept of minimizing the generator's cross-entropy instead of flipping the sign and re-design the cost function for the generator as the cross-entropy of the generated image.

\begin{equation}
J^{(G)} = -\frac{1}{2}\mathbb{E}_{z \sim p_{z}}\log D(G(z))
\end{equation}

Now the generator maximizes the discriminator being mistaken unlike previously introduced minimax game where the generator strives to minimize the discriminator being correct. This is a very heuristic method to maintain a strategy of minimizing the existing cross-entropy without a disadvantage to the generator in the actual training process. This game is no longer zero-sum game; all players have a strong gradient when the opponent is losing the game however can be considered in a cooperative relationship since each player grows further to lead growing opponent being mistaken. This is equivalent to the maximum likelihood estimation under the assumption that the discriminator is optimal. The expected gradient of this function is equal to the expected gradient of $D_{KL}(p_{data}||p_{g})$ since the problem is approximate the true data distribution by $G$. Note that minimizing KL-divergence between the training data and the model is equivalent to maximum likelihood.

To theoretically yield the global optimum of GAN, we first take the value function, $V(D,G)$ that specifies the discriminator's payoff in zero-sum game framework. Note that (3) is a heuristic mechanism to improve the actual training process. Therefore, the value function in this scenario is represented as minimization and maximization in an inner loop and outer loop, respectively. 

\begin{equation}
\min_{G} \max_{D} V(D,G) = \min_{G} \max_{D} -J^{(D)}(\theta^{(D)},\theta^{(G)})
\end{equation}

Next we take the derivative of (4) respect to a single entry $D(x)$ to obtain the optimal discriminator. In this process, the constants are ignored in advance and the expected values are formalized as integral. Let the probability distribution of real data and fake data created from the generator be denoted by $p_{data}$ and $p_{g}$ respectively. Since $G(z)$ is derived from latent variable $z$ and desired to resemble true data $x$, the cross-entropy for $G$ which is denoted by $D(G(z))$ can be re-written as $D(x)$ where $x$ is belong to $p_{g}(x)$. The optimal case for the discriminator can then be computed as: 

\begin{equation}
\max_{D} V(D,G) = \int _{x}p_{data}(x) \log D(x) + p_{g}(x)  \log (1-D(x)) dx
\end{equation}

\begin{equation}
D^{*}(x) = \frac{p_{data}(x)}{p_{data}(x)+p_{g}(x)}
\end{equation}

It is intuitively obvious that an optimal case for this scenario is $p_{g}(x) = p_{data}(x)$ because the generator creates the samples that are intended to be drawn from the same distribution as training data $x$, which would mean that the generator maximizes the discriminator being mistaken for distinction between true data $x \sim p_{data}$ and generated data $x \sim p_{g}$. Thus, the probability that the discriminator distinguishes either case is equal to $0.5$ ($D(x)=0.5$) if the generator correctly learns the distribution of true data. Note that the assumption that the discriminator is optimal is required to obtain the lower bound of this optimal case for the generator. All these can be derived by taking (6) into (5) and considering the JS-divergence (7). 

\begin{equation}
D_{JS}(p_{data}||p_{g}) = \frac{1}{2} D_{KL} \left( p_{data}||\frac{p_{data} + p_{g}}{2} \right) + \frac{1}{2} D_{KL} \left( p_{g}||\frac{p_{data} + p_{g}}{2} \right)
\end{equation}

\begin{equation}
\min_{G} V(D^{*},G) = \int _{x}p_{data}(x) \log \frac{p_{data}(x)}{p_{data}(x)+p_{g}(x)} + p_{g}(x) \log \frac{p_{g}(x)}{p_{data}(x)+p_{g}(x)} dx
\end{equation}

By solving the equivalence between (7) and (8),

\begin{equation} 
\min_{G} V(D^{*},G) = -\log(4) + 2 \cdot D_{JS}(p_{data}(x)||p_{g}(x))
\end{equation}

Finally, the optimal point for (4) is $p_{g}(x) = p_{data}(x)$ which refers to $D_{JS}(p_{data}||p_{g})=0$, hence $p_{g}(x)$ minimizing (8) has a distribution similar to $p_{data}(x)$.

Maximum likelihood estimation is the way we want to achieve high probability in all ranges where true data appears. Note that this is equivalent to minimizing cross-entropy such as (1), as described in (4). GANs are still in such estimation, however, behave in a way to get low probability in areas where true data does not appear. It shows the main difference from minimizing KL-divergence and that JS-divergence (9) is rather similar to reverse KL-divergence. The choice of divergence has not clearly explained why GANs make sharper samples, but they have received more attention as they outperform the existing generative models minimizing pixel-wise differences. 

\subsection{Image-to-Image Translation}

As previously introduced in Section 2.1, the GAN approximates the maximum likelihood using a metric of JS-divergence through sampling without explicitly defining the probability model. \cite{14} introduced GAN frameworks with the aims to obtain the generator mapping $z$ which is the latent variable, to the high dimensional space of observation $x$. Inspired by this strong ability that simply learns the distribution of $x$ by competing the generator and discriminator, compared to previous generative models, many approaches using other sources instead of $z$ that was recently proposed. 

\begin{figure}[h]
\begin{center}
\includegraphics[width=5.5cm]{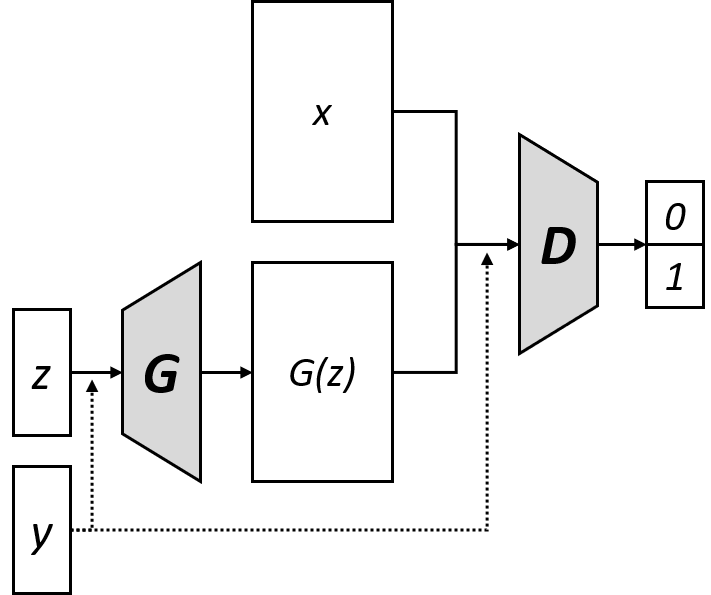}
\end{center}
\caption{The overall schematic of Conditional GANs. The key difference from the original one is conditioning the networks, in which random noise $z$ with the source data $y$ as condition is transferred to the target data domain through the generator.}
\end{figure}

They are specifically called domain-to-domain translation including text, images, audio signals and etc. with conditional probability model that generates a target when given a source. As depicted in Figure 2, it is optional to use the random noise, $z$, but the generator and discriminator's job does not change; The generator is trained to give out the output that cannot be distinguished from target images by the discriminator, which is trained to do so. Note that  most of the time, it is desirable to observe the source image $y$ for the discriminator to complete conditional probability model in adversarial training framework. Therefore, the value function in this scenario is as follows:

\begin{equation}
\min_{G} \max_{D} V(D,G) = \mathbb{E}_{x,y \sim p_{data}}\log D(x,y) + \mathbb{E}_{y \sim P_{data}, z \sim p_{z}}\log (1-D(y, G(y,z)))
\end{equation}

where $x$ is target data, and $y$ is source data according to $x$. To further improve the performance of the generator, the most common way is to use a traditional loss minimizing the distance between the source image mapped to the target domain, and its reference image, hence the model finds the properties to which they are linked between given domains providing data in pairs. 

\begin{equation}
L_{1} = \mathbb{E}_{x,y \sim p_{data}, z \sim p_{z}}||x-G(y,z)||_{1}
\end{equation}

\begin{equation}
G^{*} = \arg \min_{G} \max_{D} V(D,G) + \lambda L_{1}
\end{equation}

The generator not only fool the discriminator but also minimize L1 or L2 distance from the ground truth within pairwise data. The choice of using random noise $z$ does not significantly contribute to learning conditional probability, however the model would loss stochasticity and only produce deterministic output if $z$ is not used. This is previously employed and attempted by \cite{20, 21, 12}, but the effectiveness of random noise clearly depends on given problem type. Thus, the final objective generator of the generator is described in (12).

If pairwise data is not available, manually the feature is often determined for re-mapping to the target domain after the source is mapped to the low dimensional latent space, which suffers over-fitting. However \cite{13} proposed unpaired image-to-image translation using cycle consistency where the source image transferred to the target domain is able to be returned its original domain. This approach uses very heuristic mechanism particularly in a situation where the acquisition of pairwise data is labor-intensive, but the performance for the image quality is lower than the one that uses the pairwise data.

\section{Method}

In this chapter, we introduce our method for bone suppression using specifically designed GAN. As mentioned in previous section, the GAN approximates the intractable maximum likelihood using a metric of JS-divergence through sampling the latent variable from commonly known distribution, without explicitly defining the probability model. However, the definition of the sampling space does not fundamentally contribute to our problem since obtaining the output according to the input can be regarded as conditional image translation. A pair of the X-ray images with ribs and those with no ribs are available due to previously acquired data via DES. Therefore, L1-distance between the predicted value and the actual value for the bone suppressed image can practically guide the distribution learning with GAN. This guidance has theoretically global-convergence as the GAN approach, however, is unlikely to work a main objective function in training process. It is typically used in a weighted manner to assist the other criteria because it is one of the pixel-related functions that reduces the average difference of input and output. Here we use additional support mechanism to outperform existing state-of-the-art methods.

\subsection{Haar 2D Wavelet Decomposition}

Wavelet is a signal of the form firstly introduced by \cite{22} where a short localized oscillation repeats near zero and slowly vanishes. The wavelet is designed to have specific properties that are useful for signal processing; the convolution between wavelets and the target signal extracts certain information in a frequency or time domain. The principle can be described as the wavelet resonates if the target signal and the wavelet have the same frequency. The convolution of the signal to be analyzed with such wavelets is very similar to the Fourier Transform for examining the frequency band of a certain part of the signal. This is called wavelet transform, which is the process of separating the signal into a set of specific wavelets that are obtained from shifting or scaling one basic wavelet basis function. Its application is not only for the signal processing, but also for time series analysis or digital control system. The key features of time-frequency analysis with the wavelet transform from Short Time Fourier Transform (STFT) is that it adaptively selects frequency band based on the characteristics of the signal. The time resolution of the wavelet transform differs depending on frequency bands, whereas the STFT has same resolution at all frequency bands. Therefore, since the sudden change of the signal such as noise is very visible in frequency changing and important for perceptual quality, wavelet transform is more effective. All these performances have been verified by \cite{23, 24, 25}.

\begin{figure}[h]
\begin{center}
\includegraphics[width=11cm]{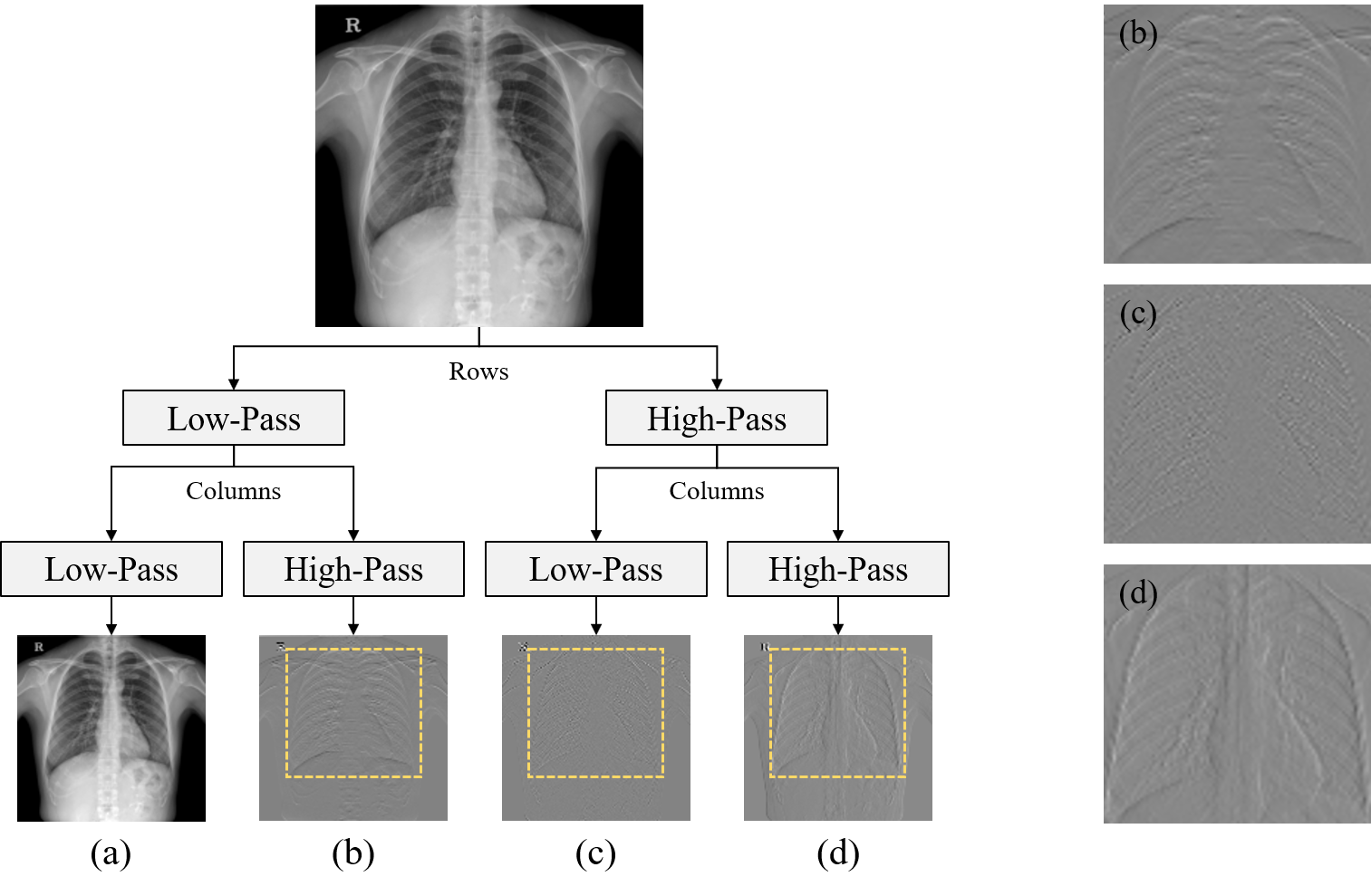}
\end{center}
\caption{Haar 2D wavelet decomposition. The row direction in image is split into high-pass and low-pass sub-bands, then the column direction repeats this step. The decomposition results are put in four components; (a) sub-sampled original image, the directional feature images in (b) vertical, (c) horizontal, and (d) diagonal details.}
\end{figure}

We adopted Haar wavelet transform, which is a one of the most popular wavelet transforms. Note that Haar wavelet is the basis wavelet in Haar wavelet transform and appears in square-shaped functions thereby is not continuous and differentiable. Haar transform using such wavelets can be used to analyze the localized feature of signals due to the orthogonal property. Our problem addresses two-dimensional signals, thus when the image is two-dimensionally wavelet-transformed, the high-frequency components are collected at the upper right and the low ones at the bottom left as shown in Figure 3. This is also regarded as 2D wavelet decomposition. 

Frequency information obtained from wavelet decomposition have a very critical role in training deep neural network. In terms of successfully applied deep learning based applications, the main strength is to approximate complex source-to-target function with non-linearity when a large scale of training data is provided. The network learns the feature of interest without manually defining the features by human that often suffer from the lack of strong prior information of source and target domain. However, directly using normal X-ray images in our case can be more challenging for the neural network. Most of the time, it is desirable to provide conceptual hints instead of entirely relying on its neural system. It also pre-defines the features that the network should learn, which allows the model to converge more quickly and efficiently. This behavior has already been proven by \cite{26} and its extension \cite{27}.

\subsection{Network Architecture}

The network architecture is based on Pix2Pix proposed by \cite{12}. The overall concept is equivalent to \cite{12}, which is that the generator minimizes pairwise difference and simultaneously attempts to fool the discriminator. In this process, GAN framework helps the network overcome the limitation by reducing the average error between input and output. In this study, we have added two purposely modified techniques to improve our specific task, bone suppression. First, as previous section introduced, we changed the input system from normal gray-scale X-ray images to wavelet decomposed X-ray images. This can efficiently decompose the directional components of X-ray, vertical, horizontal, diagonal frequency details to facilitate easier training of a deep network. Second, we partially modified training system in GAN framework, which will be further introduced in next section. The proposed model consists of the basic network in GAN; generator and discriminator. The architecture of the generator that receives the original image and produces bone suppressed images is depicted in Figure 4.

\begin{figure}[h]
\begin{center}
\includegraphics[width=15cm]{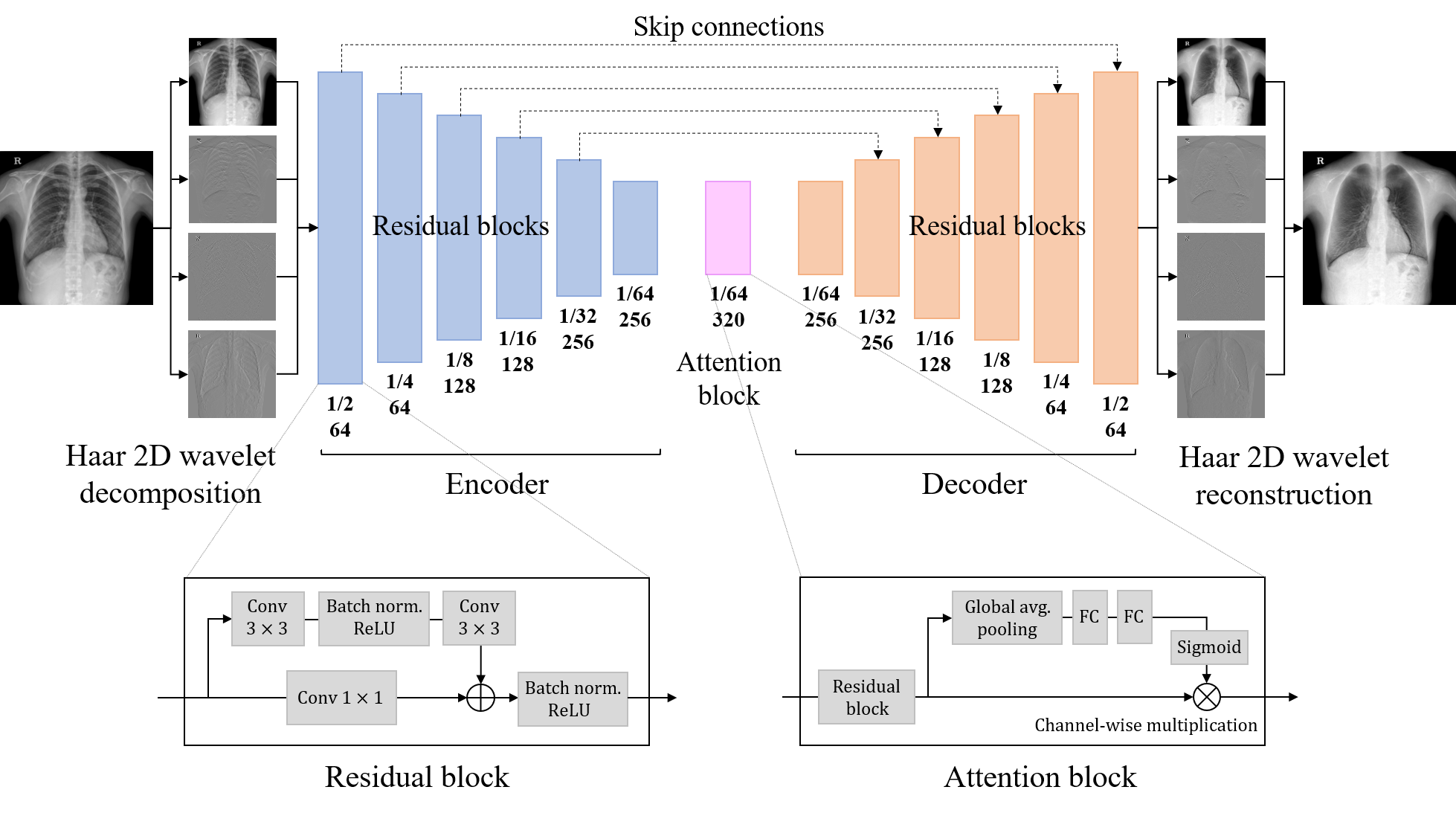}
\end{center}
\caption{The architecture of the generator. The two values below each colored block represent the sub-sampling ratio respect to the original input size, and the output channels. The residual block enhances the gradient flow of the generator by shuttling the information to the next layer, and the last encoded feature finally receives self-attention through an attention block.}
\end{figure}

The generator takes the input size as $1024\times1024$ with gray-scale (1 channel) then converts the input to $512\times512\times4$ by Haar 2D wavelet decomposition and concatenating its results. As depicted in Figure 4, the overall architecture is based on convolutional auto-encoder with skip connections, which is regarded as U-Net \cite{9}. The network consists of 12 residual blocks from \cite{28} and an attention block (a squeeze and excitation block) firstly proposed by \cite{29}. The robustness of residual network, which overcomes the limitation that deep networks are hard to train, have been proven in many computer vision tasks such as image recognition. Each residual block has two $3\times3$ convolution layers, and an additional $1\times1$ convolution layer that translates the input when changing the output channel. Translating the feature maps from shallower layer to following deeper layer has a critical role in training deep networks; it is rarely desirable for the deeper layer to directly fit the highly abstracted features, and such flow of the feature maps also improves gradient flow in back-propagation. In terms of the skip connections, the residual block in the encoder shuttles the high frequency information to its corresponding block in the decoder, thus the model can maintain the spatial frequency resolution and result in the sharp images. At the center of the network, a squeeze and excitation block is used for the attention mechanism facilitating the convergence of the model. This block summarizes all the feature maps through global average pooling, which is very important in the deep neural network where the local receptive field is small. The global spatial information is compressed into a channel descriptor and re-calibrated to calculate channel-wise dependencies.

\begin{figure}[h]
\begin{center}
\includegraphics[width=15cm]{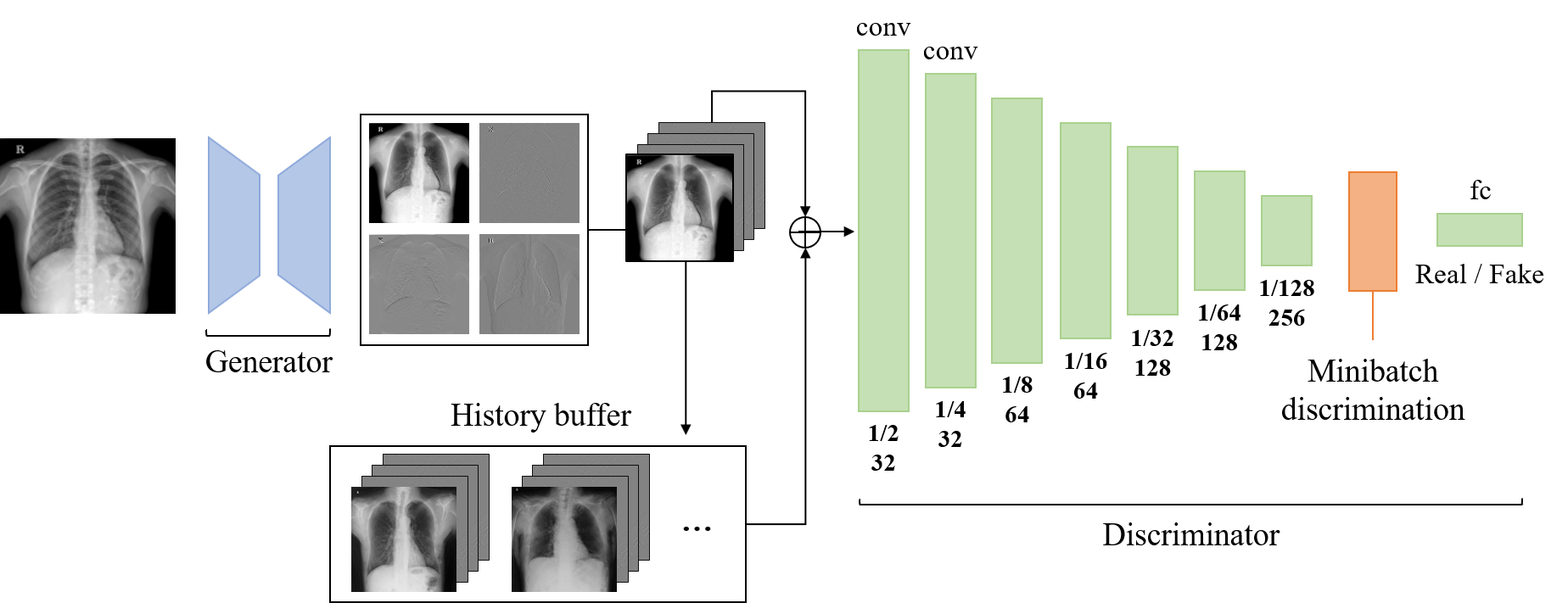}
\end{center}
\caption{The architecture of the discriminator. The numbers below each convolution block is equivalent to those in Figure 4. The discriminator also takes the history of the generator's samples and considers the distribution of batch of images instead of the single image.}
\end{figure}

The discriminator contains 7 convolution layers and a fully connected layer to output a single probability whether given image is a fake image, which comes from the generator or not. Note that a stride in convolution operation is doubled instead of using a pooling layer. Maintaining the sharpness of other tissues by removing only the ribs in X-ray corresponding to the horizontal noise is still challenging while the bone suppressed image is blurry in general convolutional auto-encoder families. In this problem, the discriminator has the most important role; the degree to which the generator gets stronger (to trick the discriminator) depends on how we design the input that the discriminator looks. Therefore, we also took four components obtained by Haar 2D wavelet decomposition as the input hence the generator not only tries to make the four components shown in Figure 3 equal to those of the output, but also simultaneously avoid the blur to fool the discriminator. To make this more useful, we added history buffer and minibatch discrimination between the last convolution layer and the fully connected layer as depicted in Figure 5, improving both discriminator and generator.

\subsection{Training}

The discriminator and generator in the proposed method models are independently parameterized, and update the parameters by stochastic gradient descent based one their objective function (to minimize the cost function). The generator optimizes the Maximum Log-Likelihood Estimation (MLE) criteria previously described in (3) and the guidance term (11) with Haar 2D wavelet decomposed details. Note that maximizing the log likelihood in the logistic regression on both discriminator and generator is equivalent to minimizing their cross entropy. The discriminator also optimizes its MLE criteria in (1). Here we use Adam optimizer \cite{32} with initial learning rate = 0.0008 and batch size = 8. 

However, the GAN still fails to fully address mode collapse although it has grown dramatically in recent years. Mode collapse is when the generator creates similar samples only where the discriminator does not distinguish well. These samples are so-called `strange' that the discriminator decided them as real and that the generator succeeds in tricking the discriminator, because such success does not consider the shape or texture that they have. This is primarily due to the loss function of the generator, which is a cross-entropy with its generated image focusing on images that are not well distinguished. In terms of adversarial frameworks, the discriminator network neither improves the generator by distinguishing all the given samples nor failing to distinguish them all, and often fail to converge. Thus, we need an equilibrium in their strength as long as using adversarial framework. In order to solve these problems and improve learning convergence speed, recurrent optimization method that involves history buffer and minibatch discrimination are used.

\subsection{History Buffer}

The history buffer is a buffer that reflects the previous training results in the next training steps by the generator saving some images it has created. The wide range occurrence of the mode collapse in training process has a critical drawback; most of deep learning frameworks that do not use recurrent network such as Long Short-Term Memory (LSTM), apply the loss and the gradient calculation only respect to the currently given batch data. For this reason, the GAN frameworks also exhibits unstable learning because the discriminator forgets the past generation. 

\begin{figure}[h]
\begin{center}
\includegraphics[width=8.5cm]{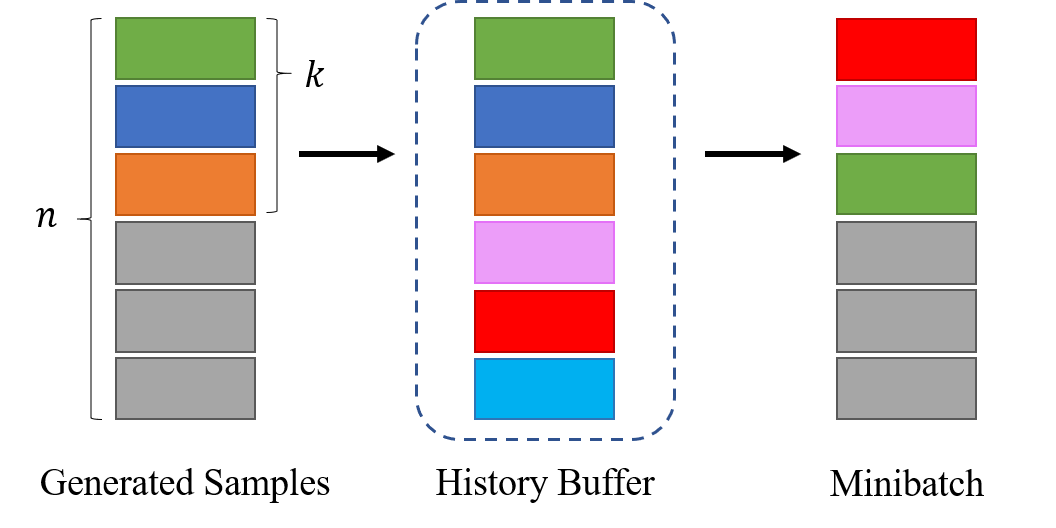}
\end{center}
\caption{The illustration of history buffer that temporarily takes the half of generated samples in minibatch, and re-fills it with the samples randomly picked after shuffling the data.}
\label{history_buffer}
\end{figure}

This problem is not first addressed in this paper, and in particular the mechanism of using the history buffer has already been proposed by \cite{30}. They noticed significant performance improvement depending on the presence of using a history of generated images. The authors of \cite{30} addressed that this lack of memory of the discriminator can cause divergence of the adversarial training, and lead the generator to re-introduce the artifacts that the discriminator had forgotten. 

The history buffer simply takes $k$ generated samples from $(x_{i1}, x_{i2}, ..., x_{ik}, x_{ik+1}, ..., x_{in})$, which is the output mini-batch in $i$-th step from the generator. Then randomly shuffling the data in the buffer, and the $k$-size of batch data in the buffer are popped and concatenated with the remaining $(x_{ik+1}, ..., x_{in})$ thereby the batch size for training the networks is constant as depicted in Figure \ref{history_buffer}. Note that the size of the history buffer is $2k$, equivalent to batch size $n$, and such concatenation is available only when the buffer is full; i.e. the initialization starts with $(x_{11}, ..., x_{1k})$, then the mini-batch in $i$-th step finally looks like $(x_{r_{1}1}, x_{r_{2}2}, ... x_{r_{n}n})$ where $r = \{ r_{1}, r_{2}, .., r_{k} \}$ is randomly picked from 1 to $i$-step. Now the Discriminator learns to distinguish all the samples from the corresponding buffer, which leads to more stable convergence of both networks and alternatively takes the same effect as recurrent optimization.

\subsection{Minibatch Discrimination}

Minibatch discrimination has been proposed by \cite{31}, which simply transposes the feature maps to measure the distance between each feature map, thereby the discriminator network sees the distribution of images in given batch instead of a single image. Mode collapse often indicates that all outputs from the generator concentrates a single data point that the discriminator currently believes is highly realistic. Setting the discriminator to identify multiple samples is a straightforward solution to address this problem. It is also regarded as exploiting the dependency among generated images in mini-batch so that the discriminator can tell the outputs of the generator to become more dissimilar to each other. 

The actual training process in an original architecture including general classification models or generative models, is to optimize the model based on the value of the objective function in mini-batch unit. Note that `mini-batch` that we typically use for gradient descent indicates the average or the sum of individually calculated for each single data. Although most of time it is preferable to observe each data independently, our main purpose of using the adversarial training framework is to emphasize the sharpness of the image. In addition, \cite{31} shows that this minibatch discrimination mechanism does not work better in the task where the goal is to obtain a strong classifier in both supervised and semi-supervised learning.

\begin{figure}[h]
\begin{center}
\includegraphics[width=7cm]{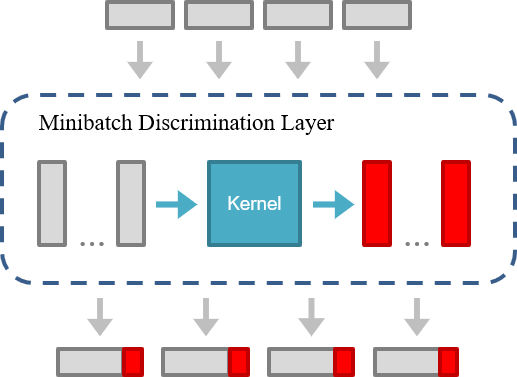}
\end{center}
\caption{The illustration of minibatch discrimination layer multiplying a specific tensor vector, measuring the distance between samples, and concatenating the results to the input.}
\label{minibatch_discrimination}
\end{figure}

Minibatch discrimination layer generally measures L1-distance between the batch of outputs that passed the last intermediate layer of the discriminator. Let the feature maps in $i$-th image in batch size of $n$ be denoted by $f(x_{i}) \in \mathbb{R}^{A}, i \in \{1, 2, ..., n\}$, where $A$ is the number of output channel. In order to get the dependency between images represented as distance, it obtains the matrix $M_{i} \in \mathbb{R}^{B \times C}$ through multiplying $f(x_{i})$ by any tensor vector (kernel) $T \in \mathbb{R}^{A \times B \times C}$ that will be optimized where $B$ and $C$ is the number of kernels and kernel size. Then it calculates the L1-distance between the rows of $M_{i,b}$ across the samples, $b \in \{1, 2, ..., B\}$ and finally applies a negative exponential $o(x_{i}) = \sum_{j=1}^{n}-e^{(||M_{i,b}-M_{j,b}||_{1})} \in \mathbb{R}^{B}$. As a results, this layer yields as many inter-dependencies among batch images as the number of kernels. The authors of \cite{31} suggest to use the other samples as `side information', thereby the output of minibatch discrimination layer is concatenated to the original feature maps on channel axis as depicted in Figure \ref{minibatch_discrimination}. The discriminator now distinguishes whether the input is a fake `batch', or a real `batch' from the training set, which allows much more visually realistic images than the one looking at a single image. 

\section{Experiment}

\subsection{Dataset}

\begin{figure}[h]
\begin{center}
\includegraphics[width=12cm]{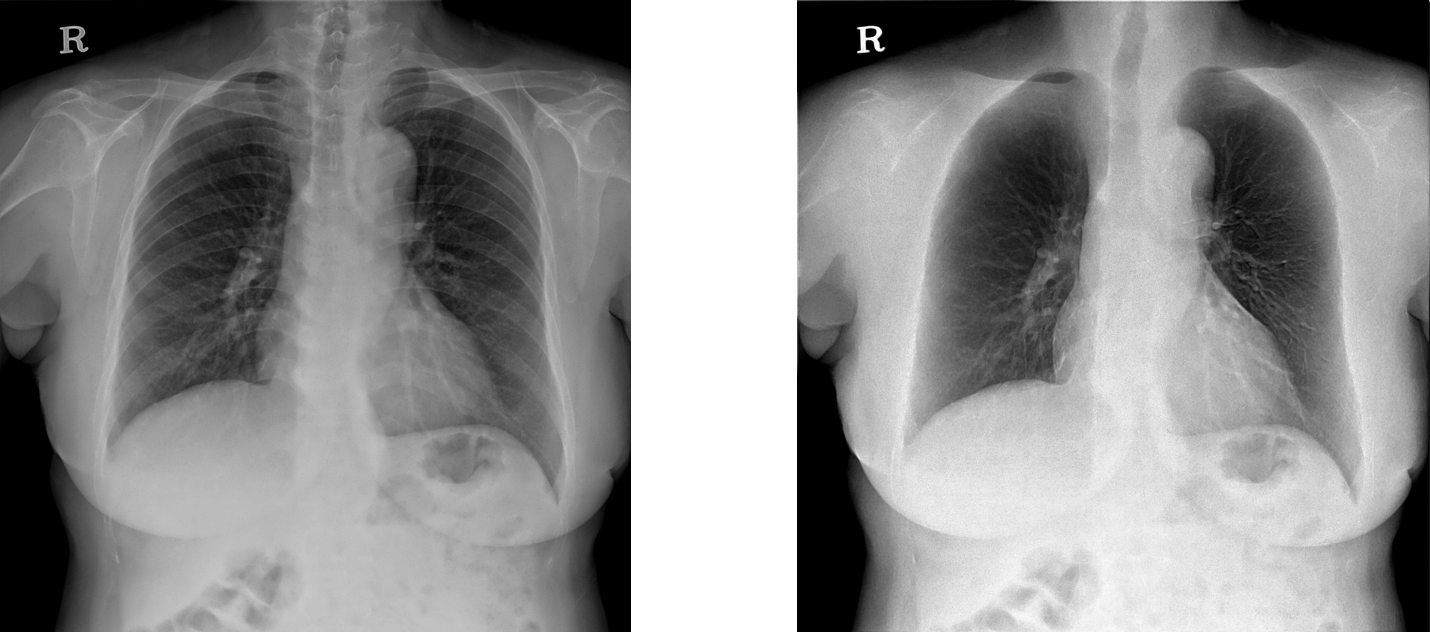}
\end{center}
\caption{Sample data of bone suppressed X-ray image via DES (right) and its original image (left).}
\label{sample_dataset}
\end{figure}

To verify the performance of the proposed model, we conducted experiments on the paired dataset of normal X-ray images and bone suppressed X-ray images via DES, which are regarded as DXRs (see Figure \ref{sample_dataset}). It contained 348 patients for paired frontal-view chest X-rays and DXRs in total, and we randomly split the dataset into 80\% for training, 10\% for validation and 10\% for test set. The dataset was originally released in DICOM format with $2017 \times 2017$ as each image size, and we rescaled them to $1024 \times 1024$ due to memory issue on GPU.

Since DICOM images exceed the commonly supported pixel dynamic range (from 0 to 255), it is preferable to select the specific dynamic range where the user tries to observe and linearly stretches the pixel intensities that lie within given range, to the original range. It is called linear windowing, and enables us to highlight bony structure rather than soft issue, or to highlight the abnormalities including lesions or at the expense of other structures present within the field-of-view. Thus, we use linear windowed images instead of a full dynamic range of images using windowing parameters provided in DICOM tags. We also normalize each image in the dataset that is subtracted by individually calculating the average of its pixels and dividing by the standard deviation.

As previously introduced in Section 1, dual energy imaging captures two radiographs at a very shot interval with different energy levels to eliminate bone by subtraction between the attenuation of soft tissue and bone at different intensities. Therefore, the artifacts may arise due to heart beat between two radiographs. We manually examined the dataset since there was no post processing to handle this problem in acquisition of original images. 11 X-ray images were excluded from the training set and used for additional test which will be discussed in Section 4.3. In addition, this paper proposes to learn bone suppression on single energy X-ray by analyzing the pair of DXRs, and we only used the X-ray images at commonly known level of energy and discard those at lower energy.

\subsection{Performance Metrics}

We consider the following three objective image quality metrics to quantitatively evaluate the proposed method. Their advantage and drawbacks outlined below:

\textbf{Peak Signal-to-Noise Ratio (PSNR)}: This metric measures the ratio between the maximum possible power of signal (pixel value) and the power of noise that corrupts the image and affects the fidelity of the image. It is an improved metric of Mean Squared Error (MSE) that does not reflect the image scale. i.e. the difference between 9 and 10 is that the pixel interval is raining from 0 to 255 (8-bit) is more noticeable than the one ranging from 0 to 4096 (12-bit). In addition, it is often expressed in logarithmic scale due to various pixel dynamic range. Given a reference $m \times n$ image $a$ and its approximation image $b$, we can obtain MSE and PSNR from the following definitions:

\begin{equation}
MSE = \frac{1}{mn} \sum_{i}^{m} \sum_{j}^{n} ||a(i,j) - b(i,j)||_{2}
\end{equation}

\begin{equation}
PSNR = 20 \log_{10}\left(\frac{\sqrt{MSE}}{MAX_{a}}\right)
\end{equation}

where $MAX_{a}$ is the maximum possible pixel value of the reference image.

\textbf{Noise Power Spectrum (NPS)} This metric gives a complete description of the noise with its amplitude over frequency resolution. It can be regarded as an improved metric of standard deviation within a specified region of interest (ROI), because the standard deviation does not consider the distribution of its noise according to frequency level. For NPS calculation, it is required to select ROI to characterizes the noise correlations with 2D Fourier Transform: 

\begin{equation}
NPS = \frac{1}{N_{ROI}} \sum_{i=1}^{N_{ROI}} \frac{1}{L_{x}L_{y}} ||FT_{2D}\{{ROI_{i}(x,y)} - \overline{ROI_{i}}\}||_{2}
\end{equation}

where $L_{x}$, $L_{y}$ are the lengths of x and y dimension of ROIs, $N_{ROI}$ is the number of ROIs used for NPS calculation, and $\overline{ROI_{i}}$ is the mean pixel value of $i$-th ROI. Note that NPS represents the noise amplitude on Fourier space in the x and y dimension, not a single value. Since the result of (15) is a spectrogram, which is a 3D figure visualized in 2D by describing the amplitude over x and y dimensional frequency with color, it is common to average this NPS along 1D radial frequency to represent spatial resolution.

\textbf{Structural Similarity Index (SSIM)}: This metric is proposed by \cite{19}, also a full reference metric such as PSNR, in which the assessment of image quality relies on an initial noise-free image. However, it improves PSNR that measures absolute pixel-by-pixel errors, considering perceptual image degradation, luminance and contrast as human-perceived change in structural information; the pixels that are spatially close are likely to have strong inter-dependencies. Given a reference image $a$ and its approximation image $b$, SSIM is defined as a product of luminance, contrast and structure functions: 

\begin{equation}
SSIM = \frac{(2\mu_{a}\mu_{b}+c_{1})(2\sigma_{ab}+c_{2})}{(\mu_{a}^{2} + \mu_{b}^{2}+c_{1})(\sigma_{a}^{2} + \sigma_{b}^{2}+c_{2})}
\end{equation}

where $\mu$ and $\sigma^{2}$ are the average and variance of corresponding image denoted by subscript, respectively. Note that $\sigma_{ab}$ is the covariance of image $a$ and $b$, and the constants $c_{1}$ and $c_{2}$ are set as $c_{1}=(0.01L)^{2}$, $c_{2}=(0.03L)^{2}$ by default where $L$ is the dynamic range of pixel.

\subsection{Quantitative results}

\begin{figure}[h]
\begin{center}
\includegraphics[width=17cm]{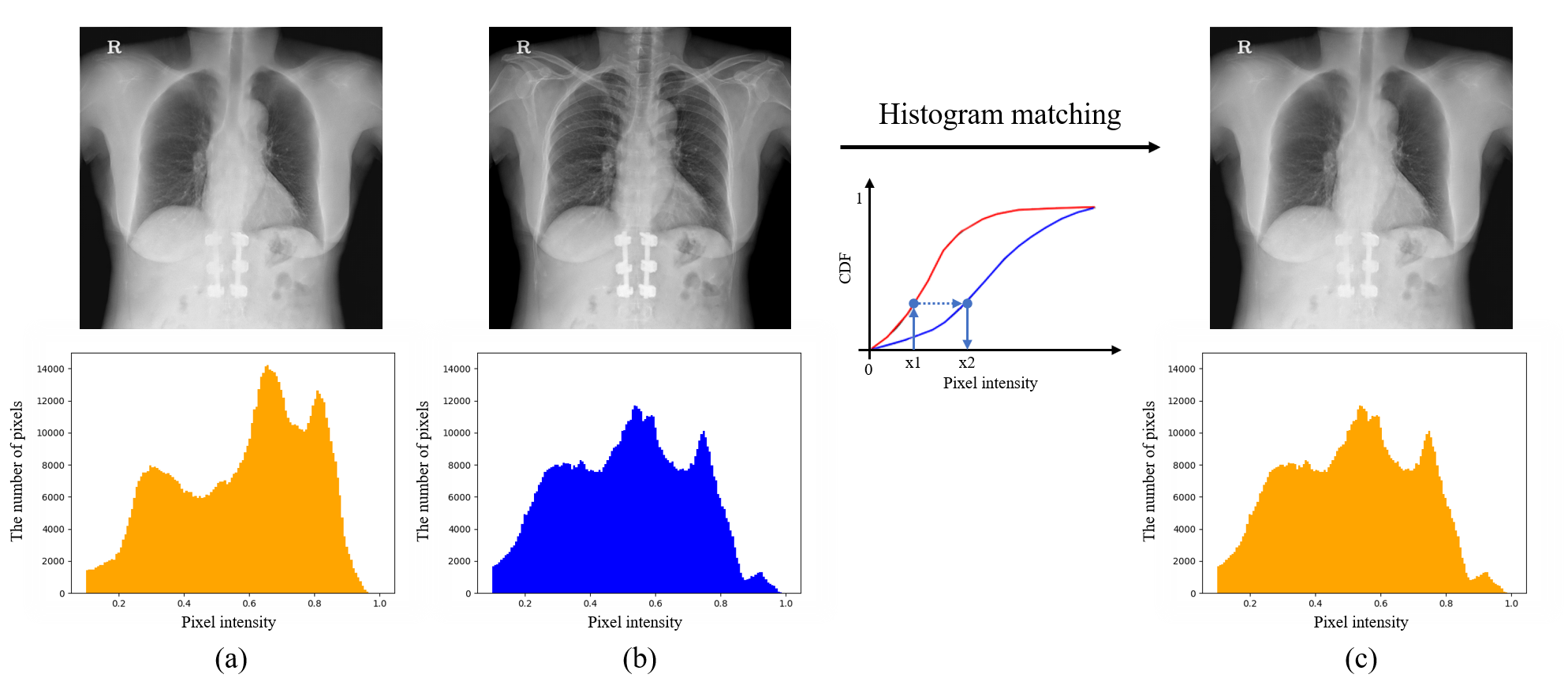}
\end{center}
\caption{How histogram matching works and the perceptual difference changes (top row) as the pixel intensities changes (bottom row): (a) target image, (b) source image and (c) histogram matched source image. Note that the DC term is omitted in each histogram.} 
\label{histogram_matching}
\end{figure}

In our overall bone suppression work-flow, we noticed the perceptual difference in the luminance due to the pixel value slightly exceeded the expected its dynamic range since there was no post-processing to adjust the pixel dynamic range of the output corresponding to its the normalized image. We could use histogram stretching, a process of simply increasing or decreasing the histogram when the images have the same contents. However, our problem takes the input as a general X-ray image and the output as a bone suppressed image. To handle this problem, we adopted histogram matching, which transforms the gray values corresponding to $i$-th cumulative histogram of the source image to have same one of the target image. 
The source image (bone suppressed image) and the target image (original image) in histogram matching are depicted in Figure \ref{histogram_matching}. Since the difference between two images was the presence of the ribs, and the pixels with the closest difference in cumulative histogram was converted first, the bone suppressed image became more visually natural; the soft tissue that appeared relatively dark due to the intensities of the bones was brightened and vice versa. Note that our initial assumption of bone suppression was not designed for musculoskeletal diagnosis and most abnormalities are more likely to be found in soft tissues with lower intensity than bones. Therefore, we concluded that histogram matching as post-processing did not severely affect the image fidelity, however in future work, we would like to further verify this issue in clinical view. 

Finally, we conducted in total three trials of training the model, and selected one model with the best performance evaluated by 34 images in validation set. Then we measured the three metrics described in previous section using the test set. The sample experiments result with the proposed method can be found in Appendix. Since the region of interest on bone suppression is lung area, the evaluation of the entire image area and the lung area is carried out. Noise Power Spectrum (NPS) is calculated by manually extracting the $120 \times 120$ ROIs for the lung area in the error (noise) matrix between the prediction and its ground truth. In addition, we proceeded simple ablation studies about how much our purposely modified technique improves the performance on bone suppression; adoption of the main network architecture as GAN and the input system as Haar 2D wavelet decomposed frequency details. The method that we propose in section 3 outperformed the rest of the differently designed models as shown in Table \ref{result-table}. 

\begin{figure}[h]
\begin{center}
\includegraphics[width=15cm]{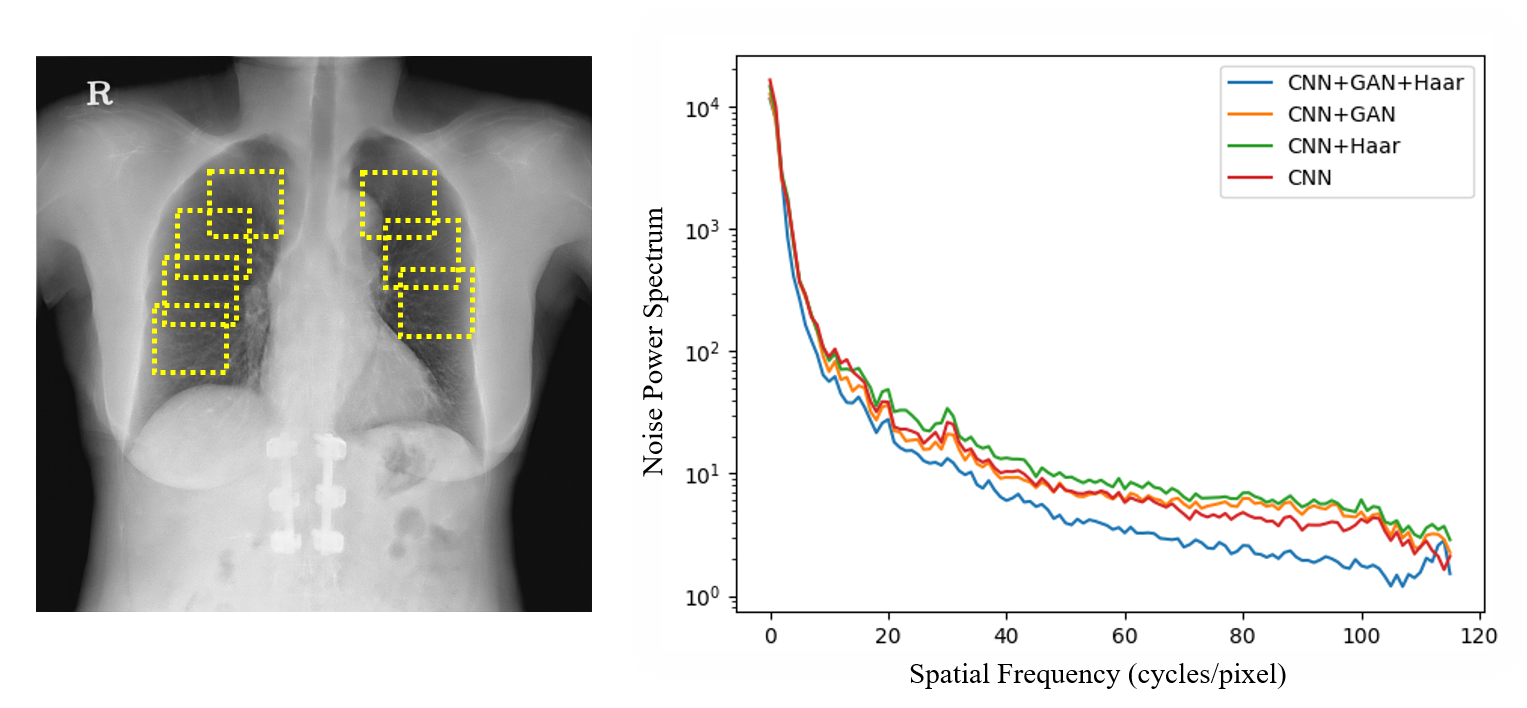}
\end{center}
\caption{Sample ROI locations (left). Only 7 ROIs are shown for clarity, but 5 $\sim$ 10 ROIs for each image are used and taken from the difference between the prediction and its ground truth. Average NPS is calculated across all patients in test set (right). }
\label{NPS}
\end{figure}

The baseline of our study, convolutional auto-encoder (CNN), has the second highest performance on both PSNR and SSIM in the lung area where as the original PSNR is low due to the overall blurry image. The CNN+Haar Wavelets shows the worst SSIM, and its bone suppressed images are very blurry and even blood vessels in the lungs are not recognizable, which will be discussed in section 4.3.2. The CNN+GAN model shows that the PSNR results are not inferior to the baseline model, however very poor SSIM results because the adversarial training sharpens the image including the bones. This may increase human-perceived changes on the ribs, which have sudden difference in the pixel intensities. Therefore, not only better removal of the bones but also high visibility due to its sharpness affects the noise power in the high frequency bands, as depicted in Figure \ref{NPS}.

\begin{table}[t]
\caption{The comparison of the performance with different conditions on the presence of purposely designed techniques in our problem.}
\label{result-table}
\begin{center}
\begin{tabular}{l|c|c|c|c}
\textbf{Model} & \textbf{PSNR} & \textbf{PSNR (Lung)} & \textbf{SSIM (Lung)} \\ \hline
CNN & 19.229 & 26.350 & 0.9031 \\
CNN + Haar Wavelets & 22.289 & 25.840 & 0.7906 \\
CNN + GAN & 21.477 & 26.343 & 0.8496 \\
\textbf{CNN + GAN + Haar Wavelets (Ours)} & \textbf{24.080} & \textbf{28.582} & \textbf{0.9304} \\
\end{tabular}
\end{center}
\end{table}

\begin{figure}[h]
\begin{center}
\includegraphics[width=16cm]{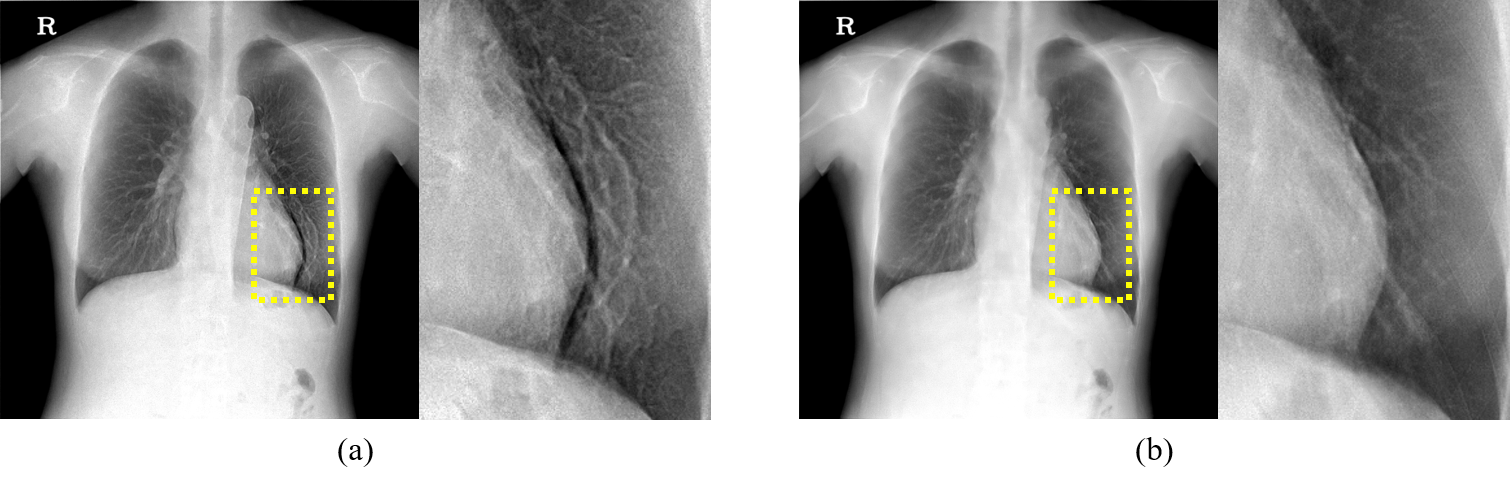}
\end{center}
\caption{The example of artifacts due to temporal interval between two radiographs in DES (a) and the results of the proposed method to first radiograph (b).}
\label{artifact}
\end{figure}

We also conducted bone suppression on the images that we manually excluded from the training set due to the conspicuous artifact. In this case, the ground truth obtained via DES can not be used as a reference image to evaluate the results. As shown in Figure \ref{artifact}, we observed, in a qualitative manner, that the motion artifacts due to heart beat did not appear and almost all information was maintained without blurry results. However, it still suffered from the lack of training data, which leads the model to often fail to capture the outline of the small blood vessels in the lungs and chest and remains further required extension of our study.

\begin{figure}[h]
\begin{center}
\includegraphics[width=16cm]{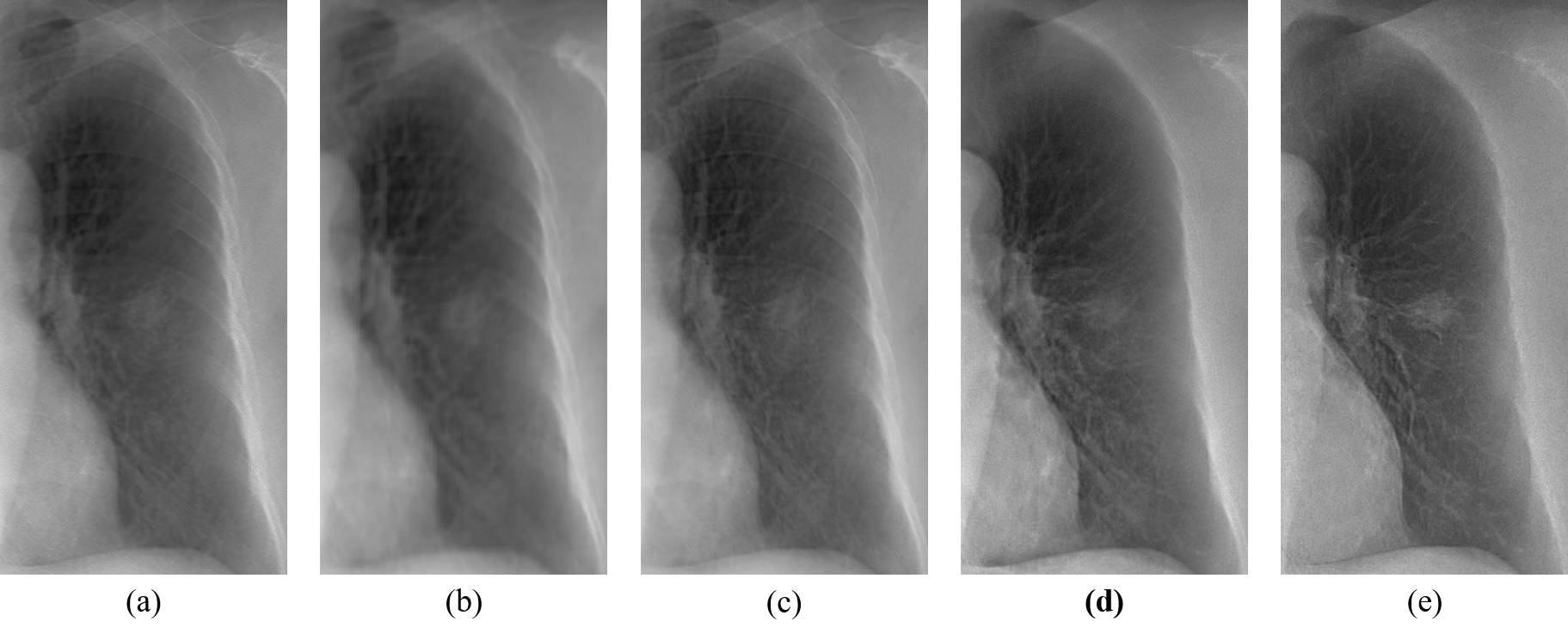}
\end{center}
\caption{The side-by-side comparison of the quality of bone suppression results with difference conditions based on the ablation studies, which is described in Table 1: (a) CNN, (b) CNN + Haar, (c) CNN + GAN, (d) CNN + GAN + Haar (ours), and (e) DES.}
\label{ablation}
\end{figure}

\subsection{Analysis of Adversarial Training}

The objective function where the discriminator distinguishes whether a given image is fake or real and the generator fools the discriminator not to do so, is very abstract. It works well even if we do not exactly define the features that we want the networks to learn in numerical form. In other words, we can only acknowledge that such features are one of style or patterns that the discriminator identifies as real. This can be solved by providing a reasonable guidance such as L1-distance to control a specific feature of interest, instead of visualizing the feature map or attention. In addition, many of GAN variants have shown sensational results beyond the pixel-related functions. When either cyclic consistency, the ability to return oneself with various domain, or the data pairs is available, it forces the training direction to make GAN converge quickly. In practice, this work verifies the quality of bone suppression using the adversarial training framework is able to outperform those with existing state-of-the-art methods.

\subsection{Analysis of Haar 2D Wavelet Decomposition}

Since our problem is de-noising the problem of considering the bone as a specific noise and removing only the bone, the bone suppression performance can be improved by providing a frequency details of the noise. Interestingly, we observed that the proposed input system, Haar 2D wavelet decomposition, works better only when used with adversarial training. As depicted in Figure \ref{ablation}, general convolutional auto-encoder with Haar wavelet decomposed information is blurrier and has less contrast. We firstly aimed to provide wavelet decomposed frequency details to help train unsupervised conditional GAN and to accelerate model convergence. However, this may act as the burden to the networks because the difference between the prediction and its ground truth becomes four times greater than the original system. When the overall data size is fixed, sharing weights for convolution for a single image is considered to be less complex compared to taking four sharing weights on each of the four images. Our proposed method specifically leverages the wavelet decomposition system and shows better results on bone suppression.

\section{Conclusion}

Bone suppression has received more attention to reduce the mis-diagnosis of radiologists due to the hidden lesion behind the bony structures. However, there are major drawbacks to currently commercialized method, dual energy subtraction (DES) within acquiring bone suppressed images. As many studies had contributed to this purpose, we successfully predicted the bone suppression results on single energy chest X-rays by analyzing previous acquired dual energy chest X-rays. We also built a model that outperforms existing approaches with a very intuitive approach; using adversarial training with frequency information as a guideline, and this method is not limited to bone suppression, but potentially contributes to other related scopes as well. Once suppressing bones on chest X-rays, the model understands the attenuation coefficient and spatial distribution of bones. In other words, it enables us to obtain that images highlighting the bony structures and bone landmarks through linear system, improving diagnosis performance on skeletal system and the registration of two chest X-rays. In future work, additional experimentation will be required to further explore the clinical meaning of this study with subjective image quality assessment.

\subsection*{Acknowledgments}

This research was supported by Basic Science Research Program through the National Research Foundation of Korea (NRF), funded by the Ministry of Education, Science, Technology (No. 2017R1A2B4004503), Hankuk University of Foreign Studies Research Fund of 2018.

\section*{Appendix}
\begin{figure}[h]
\begin{center}
\includegraphics[width=16cm]{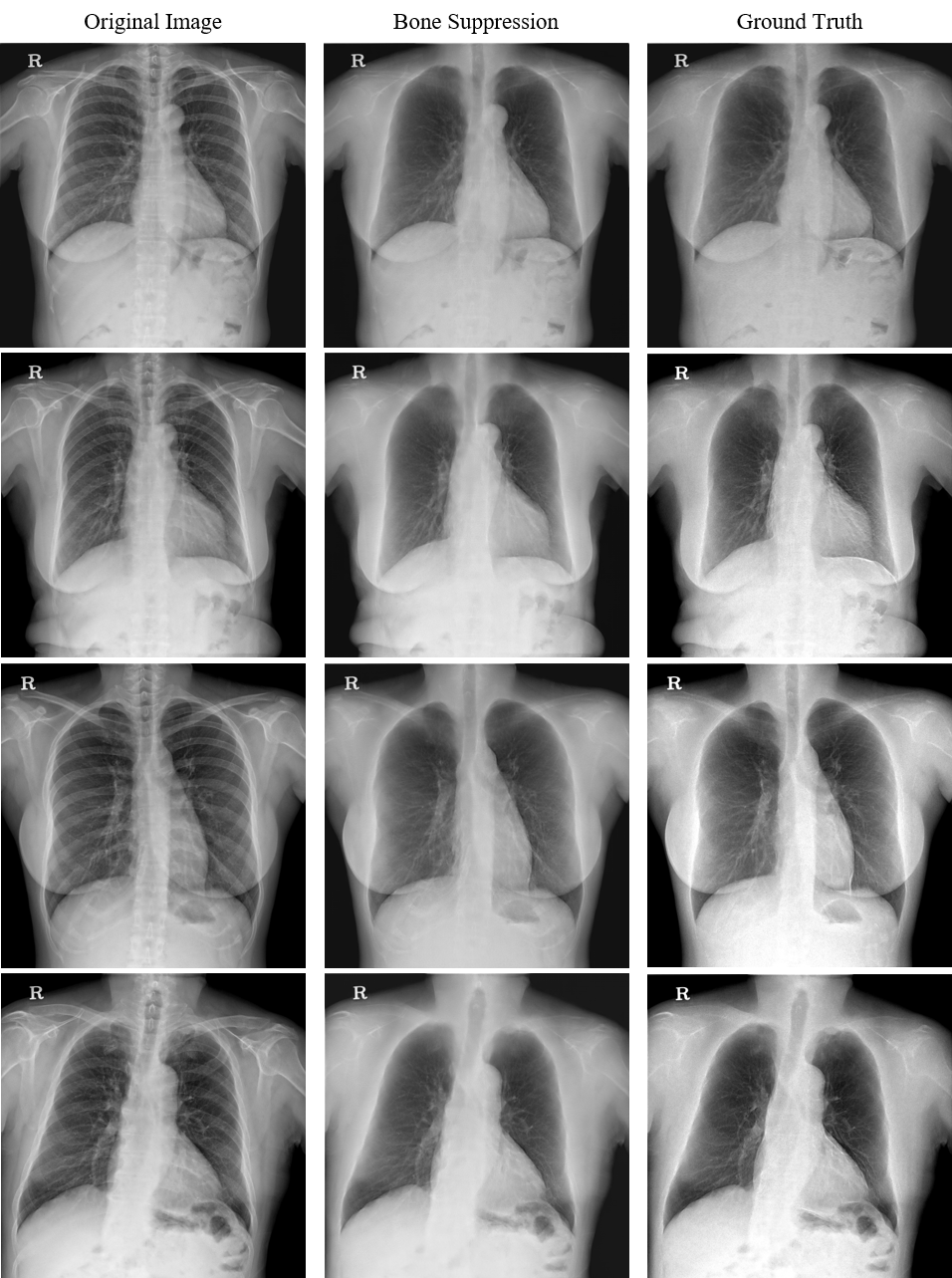}
\end{center}
\caption{The figure shows the examples of original image (right column), bone suppressed with the proposed method (center column) and ground truth obtained via DES (left column).}
\label{sample_results}
\end{figure}

We show the sample experiment results of the proposed method on single energy chest X-rays in Figure \ref{sample_results}. Note that, the original image and its ground truth in Figure \ref{sample_results} are linearly windowed using windowing parameters (default) in DICOM tags, and the bone suppressed image is histogram matched to the original one.

\end{document}